\documentclass{article}

\usepackage{microtype}
\usepackage{graphicx}
\usepackage{subfigure}
\usepackage{booktabs} % for professional tables

\usepackage{hyperref}

\usepackage[accepted]{icml2019}

\icmltitlerunning{Time-Smoothed Gradients for Online Forecasting}
\newcommand{\be}{\begin{equation}}
\newcommand{\ee}{\end{equation}}
\newcommand{\bel}{\begin{equation}}
\newcommand{\eel}{\end{equation}}
\newcommand{\bea}{\begin{eqnarray}}
\newcommand{\eea}{\end{eqnarray}}
\newcommand{\beal}{\begin{eqnarray}}
\newcommand{\eeal}{\end{eqnarray}}

\usepackage{amsmath}
\usepackage{amssymb}
\usepackage{algorithm,algorithmic}
\usepackage{mathtools}
\usepackage{multirow}
\usepackage{breqn}
\usepackage{amsthm}
\usepackage{wrapfig}
\theoremstyle{plain}
\newtheorem{thm}{Theorem}[section] % reset theorem numbering for each chapter
\theoremstyle{plain}

\theoremstyle{definition}
\newtheorem{defn}[thm]{Definition} % definition numbers are dependent on theorem numbers
\begin{document}

\twocolumn[
\icmltitle{Time-Smoothed Gradients for Online Forecasting}



\icmlsetsymbol{equal}{*}

\begin{icmlauthorlist}
\icmlauthor{Tianhao Zhu}{stevens}
\icmlauthor{Sergul Aydore}{stevens}
\end{icmlauthorlist}

\icmlaffiliation{stevens}{Stevens Institute of Technology, New Jersey, USA}

\icmlcorrespondingauthor{Sergul Aydore}{sergulaydore@gmail.com}

\icmlkeywords{Machine Learning, ICML, forecasting, online learning}

\vskip 0.3in
]

\printAffiliationsAndNotice{} % otherwise use the standard text.

\begin{abstract}
%Focus on different SGD algorithms for updating. Describe the problem as online setting. Mention the real data set. Conclude with TSSGD (talk about how less sensitive it is to the change in learning rate.).
Here, we study different update rules in stochastic gradient descent (SGD) for online forecasting problems. The selection of the learning rate parameter is critical in SGD. However, it may not be feasible to tune this parameter in online learning. Therefore, it is necessary to have an update rule that is not sensitive to the selection of the learning parameter. Inspired by the local regret metric that we introduced previously, we propose to use time-smoothed gradients within SGD update. Using the public data set-- GEFCom2014, we validate that our approach yields more stable results than the other existing approaches. Furthermore, we show that such a simple approach is computationally efficient compared to the alternatives.
\end{abstract}

\section{Introduction}
\label{introduction}
Our goal is to design efficient stochastic gradient descent (SGD) algorithms for online time-series forecasting problems. Imagine training a complex machine learning (ML) model such as recurrent neural networks (RNN). As we observe more data sets, we may need to update our model since the relationship between the inputs and the targets might change over time. In large scale ML, re-training such complex models using the entire data set will be time consuming. Ideally, we should update our model using only the new data set and automate this process. 

\citet{hazan2017efficient} introduced a notion of \textit{local regret} for online non-convex problems. They also proposed efficient algorithms that have non-linear convergence according to their proposed regret. The main idea is averaging the gradients of the most recent loss functions within a window that are evaluated at the current forecast. However, such regret definition of local regret is not suitable for forecasting problems. In forecasting, we would like to evaluate our performance on the recent loss functions that are evaluated at their corresponding forecasts instead of the current forecast.

Recently, we introduced another definition of local regret that is more interpretable for forecasting problems \cite{aydore2018local}. Under certain theoretical conditions, this regret is equivalent to the average of gradients at their corresponding forecasts over a sliding window. Inspired by this regret, we suggest using time-smoothed gradients in SGD where each gradient is computed at the corresponding forecast.

We study the stability of our approach against learning rate which is an important parameter in SGD . During online learning, tuning this parameter will not be practical. Therefore, it is important to use an algorithm which is not very sensitive to the learning rate. Moreover, an update rule in SGD should not introduce a computational bottleneck.

In this work, using a real-world time-series data set, we show that smoothing the gradients at their corresponding forecast values is an effective way for online forecasting. The advantages are: (i) it is inspired by a local regret that fits forecasting problems better, (ii) it is less sensitive to the changes in learning rate, (iii) it is faster than other alternatives.

\section{Setting}
In online forecasting, our goal is to update $x_t$ at each $t$ in order to incorporate the most recently available information. Assume that $t \in \mathcal{T} = \left\{ 1, \cdots, T \right\}$ represents a collection of $T$ consecutive points where $T$ is an integer and $t=1$ represents an initial forecast point. $f_1, \cdots, f_T : \mathcal{K} \rightarrow \mathbb{R}$ are loss functions on some convex subset $\mathcal{K} \subseteq \mathbb{R}^d$. To put in another way, $x_t$ represents the parameters of an ML model at time $t$, $f_t(x_t)$ represents the loss function computed using the available data at time $t$ given the model parameters $x_t$. In order to update $x_t$ at each $t$ using SGD, we consider the following two regret definitions.

\begin{defn}
(Hazan's local regret) The $w$-local regret of an online algorithm is defined as:
\be
HR_w(T) \triangleq \sum_{t=1}^T \| \nabla F_{t, w}(x_t) \|^2
\ee
when $\mathcal{K} = \mathbb{R}^d$ and $F_{t,w}(x_t) \triangleq \frac{1}{w} \sum_{i=0}^{w-1} f_{t-i}(x_t)$. \citet{hazan2017efficient} proposed various gradient descent algorithms where the regret $HR$ is sublinear. \label{defn:hazans_regret}
\end{defn}

\begin{defn} (Proposed Regret) We propose a $w$-local regret as:
\bea
PR_w(T) \triangleq 
%\sum_{t=1}^T \lim_{\delta \rightarrow 0} \frac{1}{\delta} \sup_{\|u\|=\delta} \left \Vert \frac{1}{w} \sum_{s=t-w+1}^t \left \alpha^s \langle  D_u(x_s), \nabla f_s(x_s) \right \rangle \right \Vert^2 
\sum_{t=1}^T  \left \Vert \nabla S_{t, w} (x_t) \right \Vert^2  
\label{eq:proposed_regret}
\eea
where $S_{t, w}(x_t) \triangleq \frac{1}{W} \sum_{i=0}^{w-1} \alpha^i f_{t-i}(x_{t-i})$, $\alpha \rightarrow 1_{-}$, $W \triangleq  \sum_{i=0}^{w-1} \alpha^i$ and $f_t(x_t) = 0$ for $t \leq 0$.
\label{defn:proposed_regret}
\end{defn}

Using our definition of regret, we effectively evaluate an online learning algorithm by computing the exponential average of gradients at the corresponding forecast values over a sliding window. This way, we assign larger weights to the most recent gradients. \citet{hazan2017efficient}'s local regret, on the other hand, computes average of previous gradients computed on the current forecast. We believe that our definition of regret is more applicable to forecasting problems as evaluating today's forecast on previous loss functions might be misleading. Algorithm \ref{alg:hazans} represents Hazan's time-smoothed SGD (HTS-SGD) algorithm which is sub-linear according to the the regret in Definition \ref{defn:hazans_regret}. Inspired by HTS-SGD, we propose time-smoothed SGD (PTS-SGD) as represented in Algorithm \ref{alg:proposed} where gradients of loss functions are calculated at their corresponding forecasts.
%\begin{algorithm}[H]
%\caption{Stochastic Gradient Descent}
%\begin{algorithmic}[1]
%  \small
%  \REQUIRE learning rate $\eta > 0$
%  \REQUIRE Set $x_1 \in \mathbb{R}^n$ arbitrarily
%
%  \FOR{$t = 1, \cdots, T$}
%       \STATE Predict $x_t$. Observe the cost function $f_t: \mathbb{R}^b \rightarrow \mathbb{R}$.
%       \STATE Update $x_{t+1} = x_t - \eta \hat{\nabla} f_{t}(x_t)$ 
%  \ENDFOR
%\end{algorithmic} \label{alg:sgd}
%\end{algorithm}%

\begin{algorithm}[H]
\caption{Hazan's Time-Smoothed Stochastic Gradient Descent (HTS-SGD)}
\begin{algorithmic}[1]
  \small
  \REQUIRE window size $w \geq 1$, learning rate $\eta > 0$
  \REQUIRE Set $x_1 \in \mathbb{R}^n$ arbitrarily

  \FOR{$t = 1, \cdots, T$}
       \STATE Predict $x_t$. Observe the cost function $f_t: \mathbb{R}^b \rightarrow \mathbb{R}$.
       \STATE Update $x_{t+1} = x_t - \frac{\eta}{w} \sum_{i=0}^{w-1} \hat{\nabla} f_{t-i}(x_t)$ 
  \ENDFOR
\end{algorithmic} \label{alg:hazans}
\end{algorithm}%

\begin{algorithm}[H]
\caption{Proposed Time-Smoothed Stochastic Gradient Descent (PTS-SGD)}
\begin{algorithmic}[1]
  \small
  \REQUIRE window size $w \geq 1$, learning rate $\eta > 0$, exponential smoothing parameter $\alpha \rightarrow 1_{-}$, normalization parameter $W \triangleq \sum_{i=0}^{w-1} \alpha^i $
  \REQUIRE Set $x_1 \in \mathbb{R}^n$ arbitrarily

  \FOR{$t = 1, \cdots, T$}
       \STATE Predict $x_t$. Observe the cost function $f_t: \mathbb{R}^b \rightarrow \mathbb{R}$.
       \STATE Update $x_{t+1} = x_t - \frac{\eta}{W} \sum_{i=0}^{w-1} \alpha^i \hat{\nabla} f_{t-i}(x_{t-i})$ 
  \ENDFOR
\end{algorithmic} \label{alg:proposed}
\end{algorithm}%
In the following sections, we study the performance of these two algorithms and standard SGD for online forecasting as well as standard SGD for offline learning. The details of these four models are described in Section \ref{sec:implementation_details}.

\section{Forecasting Overview}
Standard mean squared error as a loss function summarizes the average relationship between a set of features (regressors) and targets. The resulting forecast will be a point forecast which is the conditional mean of the value to be predicted given the input features, i.e. the most likely outcome. However, point forecasts provide only partial information about the conditional distribution of outcomes. Many business applications such as inventory planning require richer information than just the point forecasts. 

Quantile loss, on the other hand, minimizes a sum that gives asymmetric penalties for overprediction and underprediction.  For example, in demand forecasting, the penalty for overprediction and underprediction could be formulated as overage cost and opportunity cost, respectively. Hence, the loss for the ML model can be designed so that the profit is maximized. Therefore, using quantile loss as an objective function is often more desirable in forecasting applications.

The quantile loss for a given quantile $q$ between true value $y$ and the predicted value $\hat{y}$ is defined as:
\be
L_q(y, \hat{y}) = q \max(y - \hat{y}, 0) + (1 - q) \max( \hat{y} - y, 0)
\ee
where $q \in (0, 1)$.

Typically, forecasting
systems produce outputs for multiple quantiles and horizons. The total quantile loss function to be
minimized in such situations can be written as: $\sum_t \sum_k \sum_q L_q(y_{t+k}, \hat{y}^q_{t+k})$ where $\hat{y}_{t+k}^q$ is the output of the
machine learning model, e.g. RNN, to forecast the q-th quantile of horizon k at forecast creation time t.
This way, the model learns several quantiles of the conditional distribution such that $\mathbb{P}\left( y_{t+k} \leq {y}_{t+k}^q \mid y_{:t} \right) = q$.

We use quantile loss as our cost function in the following section to forecast electric demand values from a time series data set.
\section{Experimental Results}
\label{experimental result}
We conduct experiments on a real-world time series data set to evaluate the performance of our approach and compare with other SGD algorithms. We use the LSTM model since it has been shown that LSTMs are very efficient in modeling sequential data. A brief description of the data and model can be found below.
\subsection{Time Series Data set}
We use the data from GEFCom2014 \cite{barta2017gefcom} for our experiments. It is a public data set for competition in 2014. The data contains 4 sub-data sets among which we use the data that has electrical loads. The electrical load directory contains 16 sub-directories: Task1-Task15 and Solution of Task 15. Each Task1-Task15 directory contains two CSV files: benchmark.csv and train.csv. Each train.csv file contains electrical load values per hour and temperature values measured by 25 stations. The train.csv file in Task 1 contains  data from January 2005 to September 2010.  The other folders have one month of data from October 2010 to December 2012. Each benchmark.csv file  has forecasts of the electrical load values for the next month. These forecasts are generated from the benchmark method. 
\begin{figure}[!t]
    \centering 
 \includegraphics[width=0.5\textwidth]{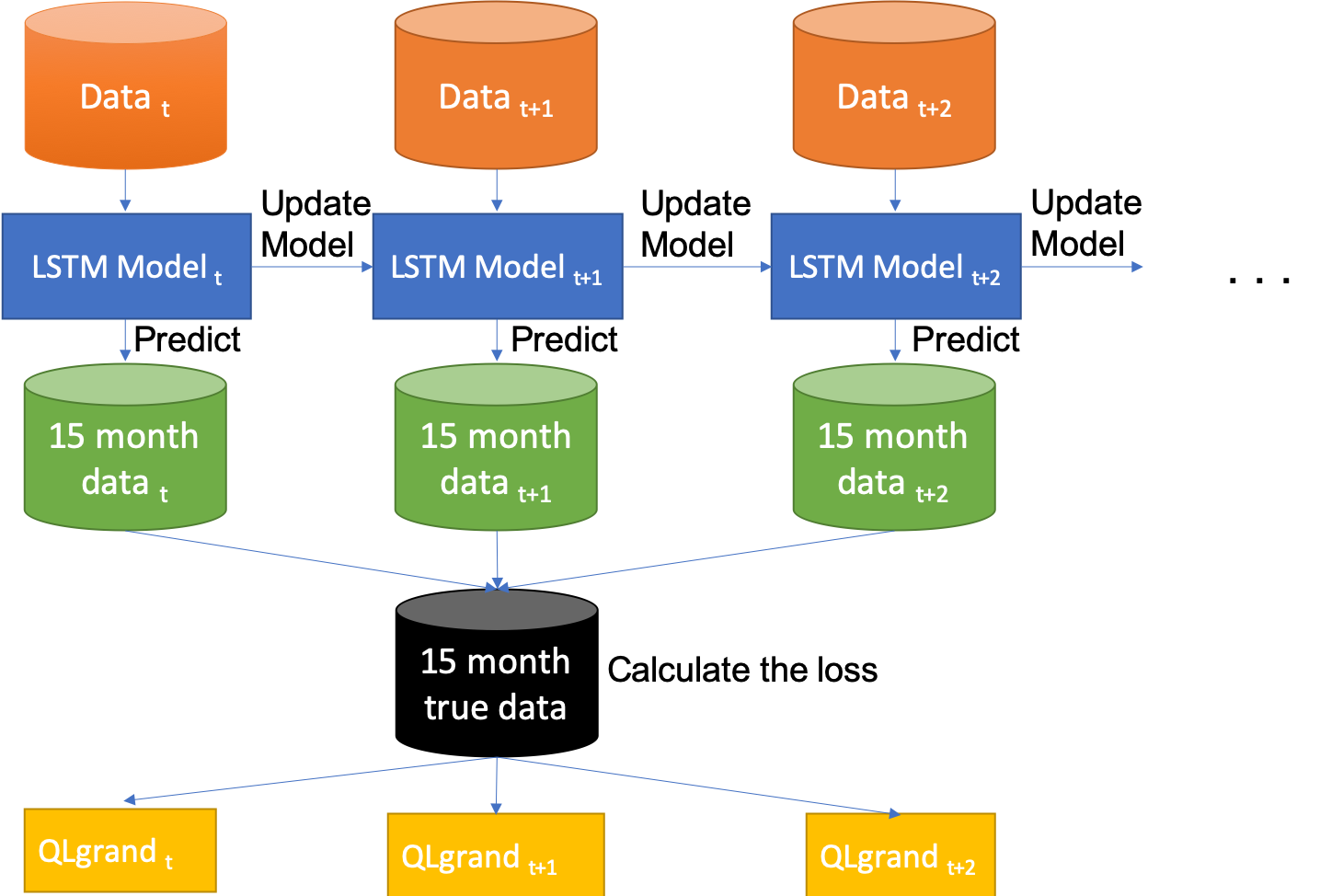} 
    \caption{The flow chart of our experiments. Each data block in orange represents a monthly data from 5-year data. The model is updated each time a monthly data arrives. Our prediction period is set to 15 months in future. Green blocks represent the forecasts for this period after each update. $QL_{grand}$ is computed using these forecasts and the true values in black.} 
    \label{Summary_Figure} 
\end{figure}
\subsection{Implementation Details}
\label{sec:implementation_details}
The general flow chart of our experiments is illustrated in Figure \ref{Summary_Figure}. We use the data from January 2005 to September 2010 for training and we set the forecast time between October 2010 and December 2012. We assume that 5-year data arrives in monthly intervals. Therefore, we update the LSTM model every time new monthly data is observed. 

\begin{figure}[!t]
    \centering 
    \includegraphics[width=0.48\textwidth]{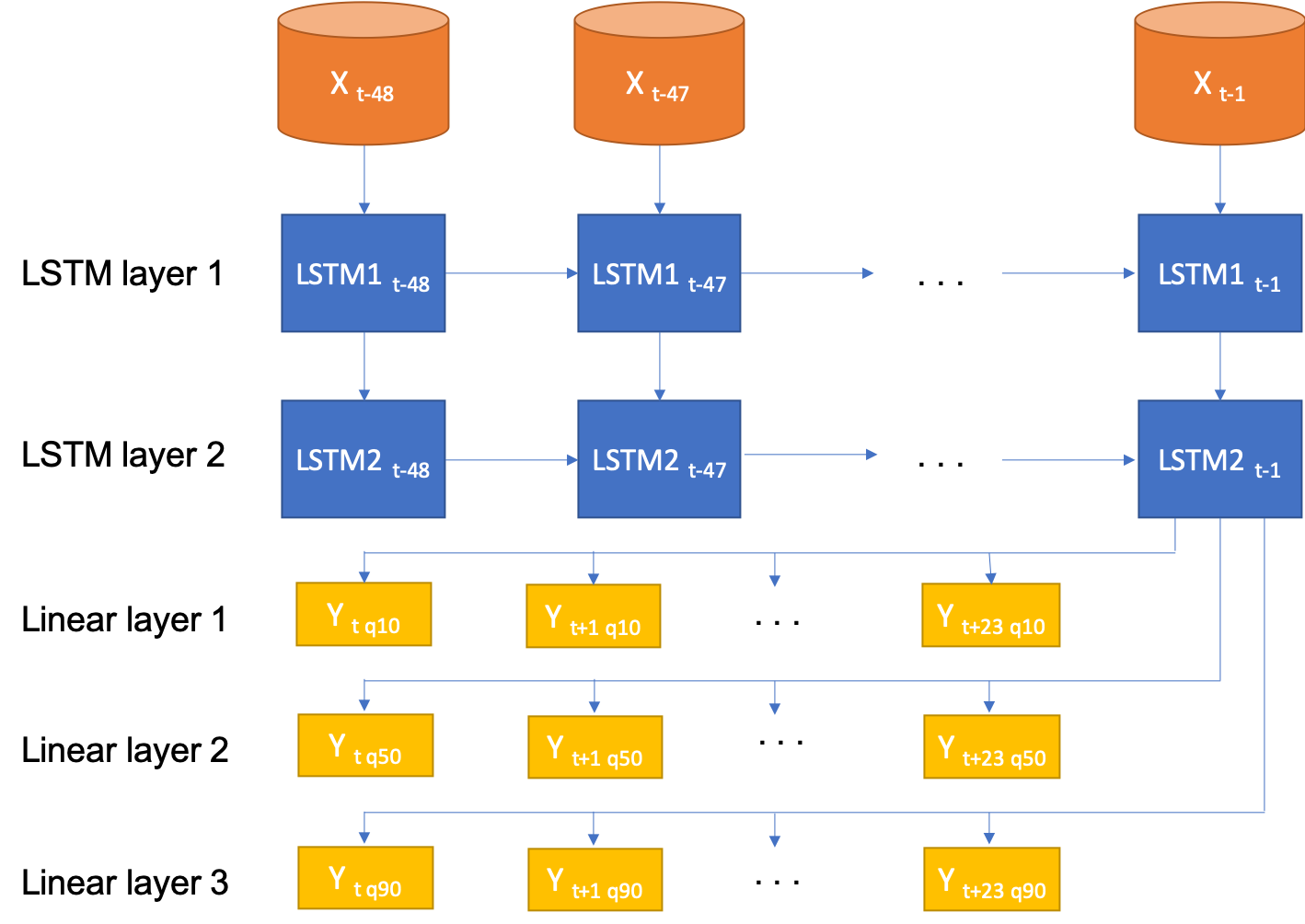} 
    \caption{The architecture of our LSTM model. We use multi-step LSTM to forecast multiple horizons. The input is two-day data of size 48 $\times$ 44 and the output is the prediction of three quantiles of  next one-day electrical load values. } 
    \label{RNN_architecture} 
\end{figure}

\textbf{LSTM Model:} LSTMs are special kind of RNNs that are developed to deal with exploding and vanishing gradient problems by introducing input, output and forget gates \cite{hochreiter1997long}. The model contains two LSTM layers and three fully connected linear layers where each represents one of the three quantiles. The architecture of our LSTM model is illustrated in Figure \ref{RNN_architecture}. We use multi-step LSTM to forecast multiple horizons. We use electrial load value, hours of the day, days of the week and months of the year as features so that the total number of features is 44. The input to our LSTM model is 48 $\times$ 44 where 48 is hours in two days. The output is the prediction of three quantiles of next day's values. 

\textbf{Training:} During the update, we make only one pass to the data, which means that the epoch number is set to $1$. In order to make learning curves smoother, we adjust the learning rate at each update $t$ so that $\eta_t \leftarrow \eta / \sqrt{t}$ where $\eta$ is the initial value for the learning rate. In our experiments, we use 1, 3, 5, 9 for the value of $\eta$.

\textbf{Metrics:} After updating the model once, we evaluate the performance on the 15 months of test data (October 2010 - December 2012). We compute quantile loss for each month and report the average of these which we call $QL_{grand}$. Low $QL_{grand}$ indicates better performance.

\begin{figure*}[!t]
    \centering 
    \hspace{-0.6cm}
    \subfigure[$\eta$=1]{
    \label{QL_lr=1}
    \includegraphics[width=0.26\textwidth]{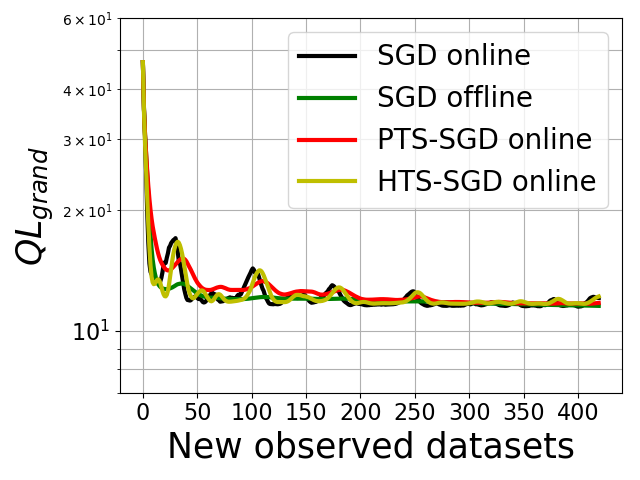}}
        \hspace{-0.35cm}
    \subfigure[$\eta$=3]{
    \label{QL_lr=3}
    \includegraphics[width=0.26\textwidth]{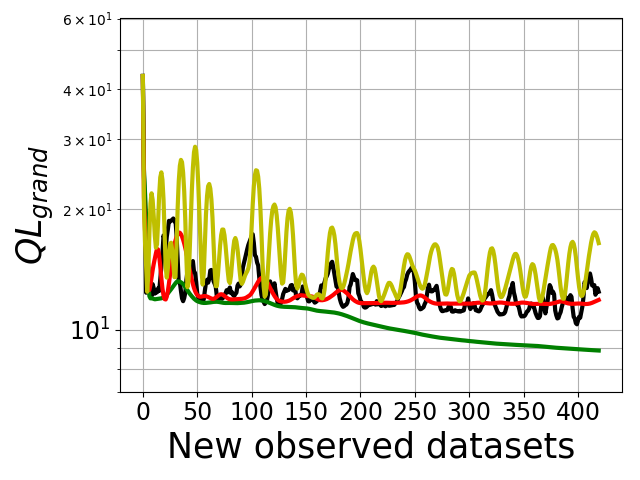}}
            \hspace{-0.35cm}
    \subfigure[$\eta$=5]{
    \label{QL_lr=5}
    \includegraphics[width=0.26\textwidth]{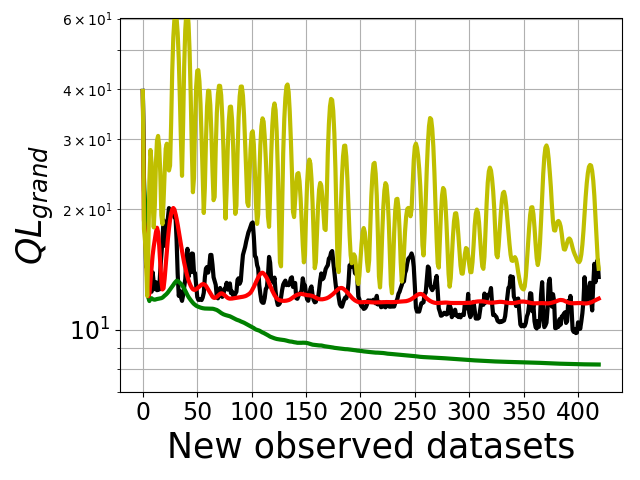}}
        \hspace{-0.35cm}
    \subfigure[$\eta$=9]{
    \label{QL_lr=9}
    \includegraphics[width=0.26\textwidth]{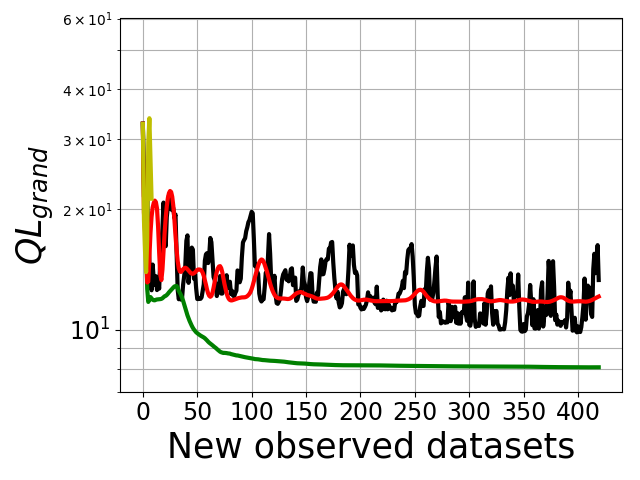}}
\caption{Comparison of models in terms of accuracy for various learning rates. PTS-SGD is less sensitive to $\eta$ than SGD online and HTS-SGD. SGD offline performs the best as expected and yields higher accuracy as $\eta$ increases.}
\label{Comparison QL{grand}}
\end{figure*}

\textbf{Methods}: We use one offline and three online methods for training. Offline model uses standard SGD algorithm and is re-trained from scratch to incorporate all the available data each time new data arrives. We see this strategy as the best strategy to be achieved, but as the most expensive  in terms of computation. We call this SGD offline in our experiments. The online models are updated only once each time new data is observed.  We use standard SGD (called SGD online), Hazan's time smoothed SGD (called HTS-SGD) and our proposed time smoothed SGD (called PTS-SGD) for online models.

\textbf{Computational Details:} We use Python 3.7 for implementation \cite{oliphant2007python} using open source library PyTorch \cite{paszke2017pytorch}. We use 2 NVIDIA GeForce RTX 2080 Ti GPUs with 512 GB Memory to run our experiments.
\begin{figure*}[!t]
    \centering 
    \hspace{-0.61cm}
    \subfigure[$w$=20]{
    \label{time_w=20}
    \includegraphics[width=0.255\textwidth]{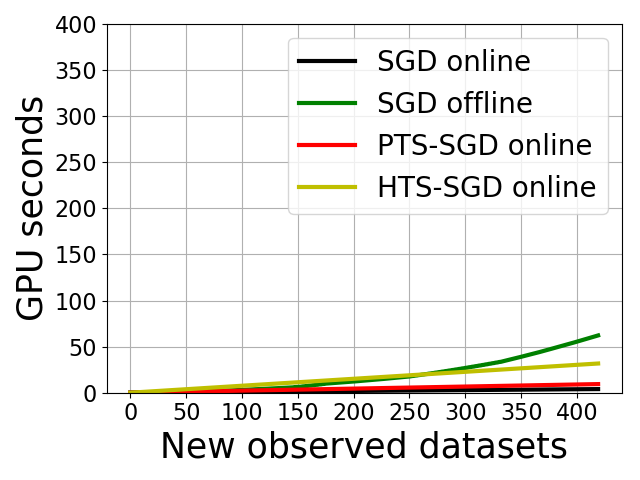}}
        \hspace{-0.35cm}
    \subfigure[$w$=50]{
    \label{time_w=50}
    \includegraphics[width=0.255\textwidth]{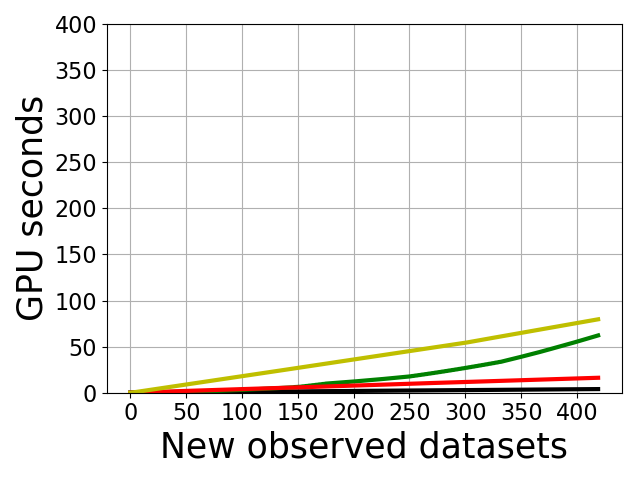}}
            \hspace{-0.35cm}
    \subfigure[$w$=150]{
    \label{time_w=150}
    \includegraphics[width=0.255\textwidth]{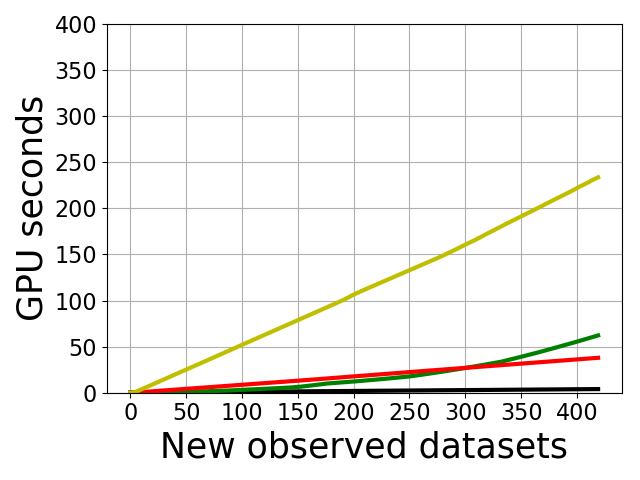}}
        \hspace{-0.35cm}
    \subfigure[$w$=200]{
    \label{time_w=200}
    \includegraphics[width=0.255\textwidth]{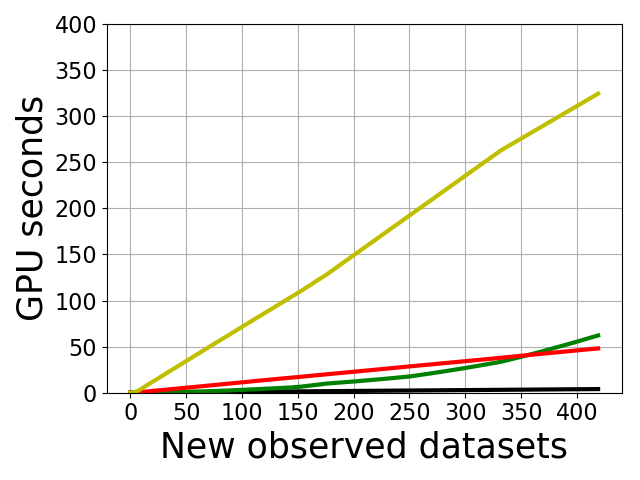}
   }
\caption{Comparison of computation time between four models  with varying $w$ when $\eta=9$. Computation time for
HTS-SGD and PTS-SGD increases as $w$ increases. PTS-SGD eventually becomes more efficient than the benchmark SGD offline even for large $w$.}
\label{Comparison times}
\end{figure*}

\section{Results}
We compare the performance of three online models in terms of their (i) stability against the selection of learning rate, and (ii) computational efficiency. 

\noindent
\textbf{Stability Against Learning Rate:}
Figure \ref{Comparison QL{grand}} shows the comparison of models in terms of $QL_{grand}$ for different learning rates. When the learning rate is $1$ (Figure \ref{QL_lr=1}), the performance of all four approaches are similar. However, we would expect SGD offline model to perform the best. The results in Figure \ref{QL_lr=1} indicate that SGD offline has not converged yet and we need to use a different learning rate. We plot the performances for larger learning rates in Figures \ref{QL_lr=3}, \ref{QL_lr=5} and \ref{QL_lr=9}. The results show that larger learning rate is needed for SGD offline and it is the best performing model as expected. However, the results for SGD online and HTS-SGD oscillate a lot indicating that they are very sensitive to the changes in learning rate. Our proposed approach PTS-SGD, on the other hand, stays robust as we increase the learning rate. Note that, for $\eta=9$, the values for HTS-SGD became \textit{nan} (not a number) due to very large losses after some number of iterations, hence are not shown in the Figure.

\noindent
\textbf{Computation Time:}
We further investigate the computation time of each model. Figure \ref{Comparison times} shows the amount of time spent in terms of GPU seconds at each update for $\eta=9$ and varying $w$ for HTS-SGD and PTS-SGD. Note that, these results will not be different for other learning rates since computation time does not depend on the learning rate. The figure shows that the elapsed time increases for HTS-SGD and PTS-SGD as $w$ increases as expected. When $w=200$, HTS-SGD takes the most time. However, it can be seen that the time elapsed curve for SGD offline looks more exponential than linear. This means that at some point in future, HTS-SGD will be more efficient than SGD offline. The computation time for our PTS-SGD is already more efficient than SGD offline after 350-th observation even when $w=200$. The reason why HTS-SGD is not as efficient as PTS-SGD is because it needs to store previous losses and compute the gradients using the current parameters. Unsurprisingly, SGD online is the most efficient but its accuracy results in Figure \ref{Comparison QL{grand}} were not as stable as that of PTS-SGD.

\section{Conclusion}
In this work, we propose exponentially time-smoothed gradient descent  for online forecasting. Our approach is inspired by the regret that is more applicable for forecasting problems. The main idea is to smooth the gradients in time when an update is performed using the new data set. We evaluate the performance of this approach compared to the existing approaches as well as the  offline model. We use a real-world data set to compare all models in terms of  computation time and stability against learning rate. Our results show that our proposed algorithm (PTS-SGD) has the following benefits: (i) it is not sensitive to the learning rate, and (ii) it is computationally efficient compared to the alternatives. We believe that our contribution can have a significant impact on applications for online forecasting problems.

\bibliography{mybib}
\bibliographystyle{icml2019}

\end{document}